\documentclass{article}

\usepackage{arxiv}
\usepackage{setspace}
\usepackage[utf8]{inputenc} 
\usepackage[T1]{fontenc}    
\usepackage{hyperref}       
\usepackage{url}            
\usepackage{booktabs}       
\usepackage{amsfonts}       
\usepackage{nicefrac}       
\usepackage{microtype}      
\usepackage{cleveref}       
\usepackage{lipsum}         
\usepackage{graphicx}
\usepackage{natbib}
\usepackage{doi}
\usepackage{verbatim}
\usepackage{multirow}
\usepackage{booktabs}

\title{MobileFlow: A Multimodal LLM for Mobile GUI Agent}
\date{}

\newif\ifuniqueAffiliation
\uniqueAffiliationtrue
\ifuniqueAffiliation 
\author{ Songqin Nong\thanks{Equal Contribution}, Jiali Zhu\footnotemark[1], Rui Wu\footnotemark[1], Jiongchao Jin, Shuo Shan, Xiutian Huang, Wenhao Xu \\
\\
Ant Group\\
\\
\texttt{\{nongsongqin.nsq, zhujiali.zjl, guli.wr, jinjiongchao.jjc\}@antgroup.com}\\
\texttt{\{shanshuo.ss, huangxiutian.hxt, hao.xuwh\}@antgroup.com}
}
\else
\usepackage{authblk}

\newbox{\orcid}\sbox{\orcid}{\includegraphics[scale=0.06]{orcid.pdf}} 
\author[1]{%
	\href{https://orcid.org/0000-0000-0000-0000}{\usebox{\orcid}\hspace{1mm}David S.~Hippocampus\thanks{\texttt{hippo@cs.cranberry-lemon.edu}}}%
}
\author[1,2]{%
	\href{https://orcid.org/0000-0000-0000-0000}{\usebox{\orcid}\hspace{1mm}Elias D.~Striatum\thanks{\texttt{stariate@ee.mount-sheikh.edu}}}%
}
\affil[1]{Department of Computer Science, Cranberry-Lemon University, Pittsburgh, PA 15213}
\affil[2]{Department of Electrical Engineering, Mount-Sheikh University, Santa Narimana, Levand}
\fi

\hypersetup{
pdftitle={A template for the arxiv style},
pdfsubject={q-bio.NC, q-bio.QM},
pdfauthor={David S.~Hippocampus, Elias D.~Striatum},
pdfkeywords={First keyword, Second keyword, More},
}

\begin{document}

\maketitle
\begin{abstract}
The ongoing evolution of multimodal large-scale models, such as GPT-4v, Qwen-VL-Max, has significantly bolstered the capabilities of image comprehension and user action analysis, showcasing the potentiality of intelligent graphically-oriented user interface (GUI) assistants. However, current GUI Agents often need to access page layout information through calling system APIs, which may pose privacy risks, and also need to fix user interfaces to a certain low resolution might result in the loss of fine-grained image details. Meanwhile, the multimodal large models built for GUI Agents currently have poor understanding and decision-making performance when dealing with Mandarin apps. This paper introduces MobileFlow, a multimodal large language model meticulously crafted for mobile GUI agents. Transforming from the open-source model Qwen-VL-Chat into GUI domain, MobileFlow contains approximately 21 billion parameters and is equipped with novel hybrid visual encoders, making it possible for variable resolutions of image inputs and good support for multilingual GUI. By incorporating Mixture of Experts (MoE) expansions and pioneering alignment training strategies, MobileFlow has the capacity to fully interpret image data and comprehend user instructions for GUI interaction tasks. Finally, MobileFlow outperforms Qwen-VL-Max and GPT-4v in terms of task execution by GUI agents on both public and our proposed evaluation metrics, and has been successfully deployed in real-world business contexts, proving its effectiveness for practical applications.
\end{abstract}

\section{Introduction}
Large Language Models (LLMs) have contributed significantly to the advancement of Artificial General Intelligence (AGI) systems, demonstrating exceptional capabilities in handling human-like interaction tasks. The progress of LLMs has also led to substantial breakthroughs in Multimodal Large Language Models (MLLMs) (\cite{chen2024internvlscalingvisionfoundation, liu2024improvedbaselinesvisualinstruction, liu2023visualinstructiontuning, zhu2023minigpt4enhancingvisionlanguageunderstanding}), facilitating complex visual-language dialogue and interaction, and bridging the gap between textual and visual information. This has created a favorable opportunity for developing autonomous, GUI agents in digital worlds.
\begin{figure}[htbp]  
    \centering  
    \includegraphics[width=0.85\textwidth]{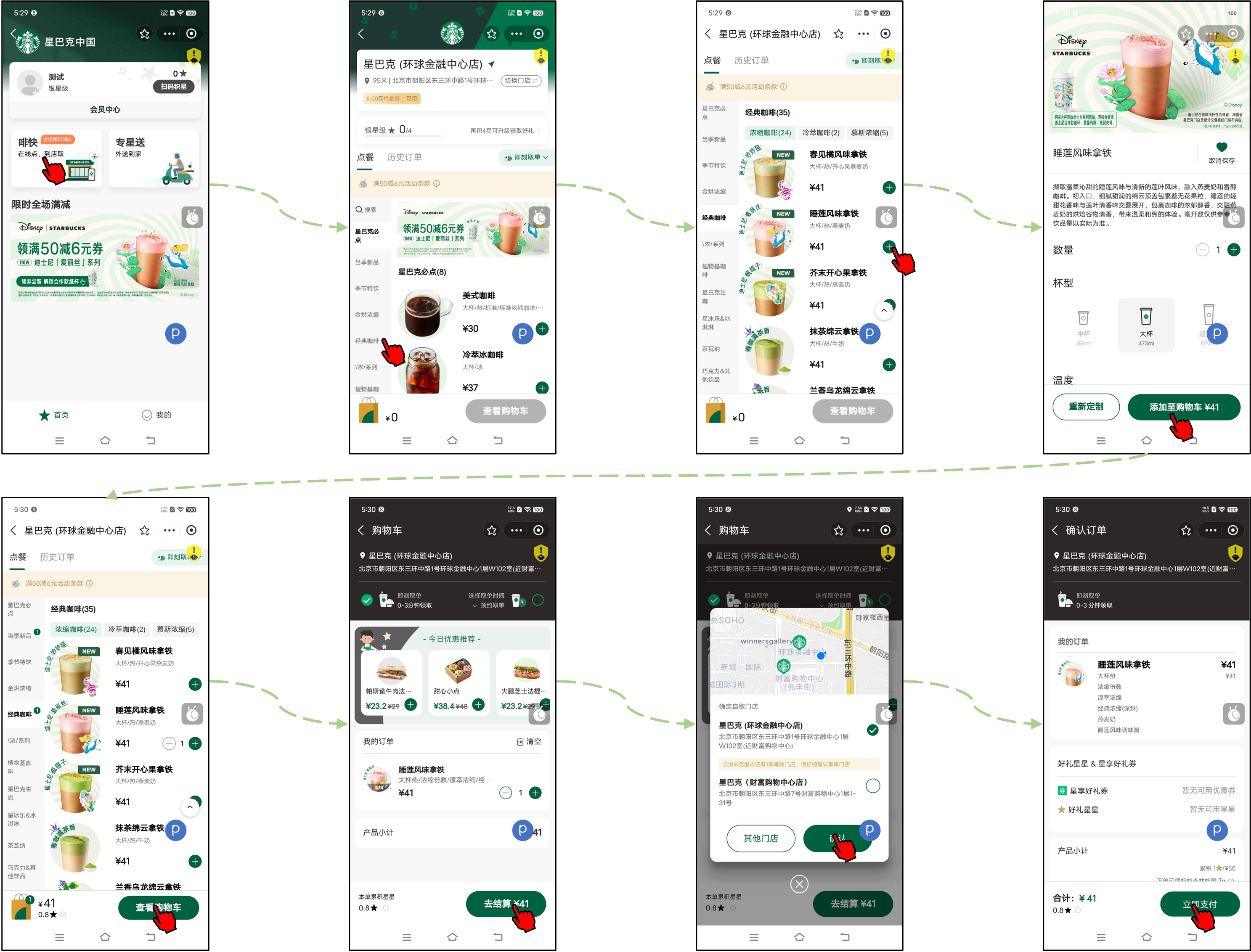}  
    \caption{Showcase of MobileFlow's application for GUI Agent. User's instruction: Get me a cup of lily latte in classic coffee at Starbucks, and I want to pick it up at store.}  
    \label{fig:figure2}  
    \vspace{-1em}
\end{figure}

Visually-enabled agents have immense potential in the real world, as they can directly perceive visual signals and interact with humans and GUIs. Vision-Language Models (VLMs) can acquire skills such as reading and programming, further expanding their potential with multi-modal information. Some prior research has started to utilize VLM models to achieve universality in GUI tasks. Agents like AppAgent (\cite{zhang2023appagentmultimodalagentssmartphone}), CogAgent (\cite{hong2023cogagentvisuallanguagemodel}), and MobileAgent (\cite{wang2024mobileagentautonomousmultimodalmobile}) have extended the reach of multimodal capabilities to GUI interfaces, representing a significant stride toward the realization of practical visual-language intelligent assistants.

Nevertheless, Multimodal agents using GPT-4v face issues with Mandarin text in GUIs, compounded by system API invocation, HTML parsing, and privacy concerns. GUIs' diverse elements challenge current agents, often using CLIP for pretraining on natural scenes(e.g., flickr30k (\cite{plummer2016flickr30kentitiescollectingregiontophrase})), insufficient for UI image text and layout extraction. VLMs' visual encoders are limited by fixed image resolutions, impacting performance in diverse "super-app" GUI scenarios. Thus, in this paper, we propose MobileFlow, a novel multi-modal Large Language Model (LLM) specifically designed for GUI Agents, standing out for its proficiency in managing applications that feature extensive Mandarin content, such as Alipay. A key component of MobileFlow is its hybrid visual encoder, which has been rigorously trained on a vast array of GUI pages. This extensive training enables MobileFlow to effectively extract and comprehend information across diverse GUI interfaces. By relying on a purely visual perception approach, MobileFlow's GUI agent eliminates the need to access system APIs to obtain page layout details. This approach not only streamlines the process but also mitigates the risk of privacy intrusion on user devices. 

Besides, MobileFlow excels in understanding GUI info, offering step-by-step user guidance, and performing GUI-specific info extraction and QA. It integrates visual and textual data via MoE and specialized GUI training. A CoT approach during fine-tuning enhances accuracy by showing reasoning, making MobileFlow effective in real-world business applications.

\subsection{Related Work}

\label{sec:related work}

\textbf{Visual Language Models}
Visual Language Large models usually consist of three parts. An encoder projects various modalities into a high-dimensional space, followed by a module that aligns the information from all modalities within this space, and finally, a decoder interprets the aligned information back into a specific modality. And as for implementation, VLMs can be divided into two types: one employs dedicated visual encoders like ViT (\cite{dosovitskiy2021imageworth16x16words}), specific alignment modules like Qformer (\cite{li2023blip2bootstrappinglanguageimagepretraining}) (or MLP), combined with a trained LLM to form a system, such as LLAVA, MiniGPT-4, Qwen-VL (\cite{bai2023qwenvlversatilevisionlanguagemodel}), CogView (\cite{ding2021cogviewmasteringtexttoimagegeneration}), etc.; the other type eliminates the independent visual and alignment modules, converting visual and textual content into tokens that are then fed directly into a Decoder-only architecture LLM, creating a unified end-to-end VLM, such as Fuyu-8B (\cite{fuyu-8b}), Chameleon (\cite{chameleonteam2024chameleonmixedmodalearlyfusionfoundation}), and so on. 

\textbf{GUI Agents}
CogAgent is constructed based on CogVLM (\cite{wang2024cogvlmvisualexpertpretrained}) and relies solely on image information, avoiding the need for system API calls, but its LLM component has not undergone targeted training. MobileAgent and AppAgent utilize existing VLMs like GPT-4v to construct UI Agents, leaning more towards prompt engineering while also depending on external modules or APIs to obtain information under the UI layer. Ferret-UI (\cite{you2024ferretuigroundedmobileui}), derived from Ferret (\cite{you2023ferretrefergroundgranularity}), supports arbitrary resolution image input, yet it still follows the traditional training approach for dialogue and question-answering in general scenarios without optimizing for UI Navigation capabilities.

\textbf{Vision Foundation Models for VLMs}
For VLMs, the visual encoder component is of paramount importance as it determines what can be "seen". Notably, widely-utilized architectures such as CLIP-ViT (\cite{radford2021learningtransferablevisualmodels}) and SigLIP (\cite{zhai2023sigmoidlosslanguageimage}) have spurred a series of studies aimed at identifying the most suitable visual encoders for integration into VLMs. For instance, identified marked differences in the visual representations captured by CLIP and DINOv2 (\cite{oquab2024dinov2learningrobustvisual}), leading to the creation of a mixed module that integrates features from both models. Moreover, different approaches have been introduced that employ varied visual encoders to process images at distinct resolutions, thereby combining features at various levels of abstraction. For example, LLaVA-HR (\cite{liu2023visualinstructiontuning}) features a bifurcated visual encoder that combines CLIP-ViT with CLIP-ConvNext[20], while DeepSeek-VL (\cite{deepseekai2024deepseekllmscalingopensource}) incorporates SigLIP-L and SAM-B. These methodologies consistently leverage pre-trained visual perceptors. In this research, we introduce LayoutLMV3 (\cite{huang2022layoutlmv3pretrainingdocumentai}), a visual branch pretrained on extensive UI data, capable of dynamically adjusting to UI images with varying aspect ratios, thereby enhancing the understanding capabilities of the overall visual encoder and its applicability within GUI agents.

\section{Method}

\begin{figure}[htbp]  
    \centering  
    \includegraphics[width=0.9\textwidth]{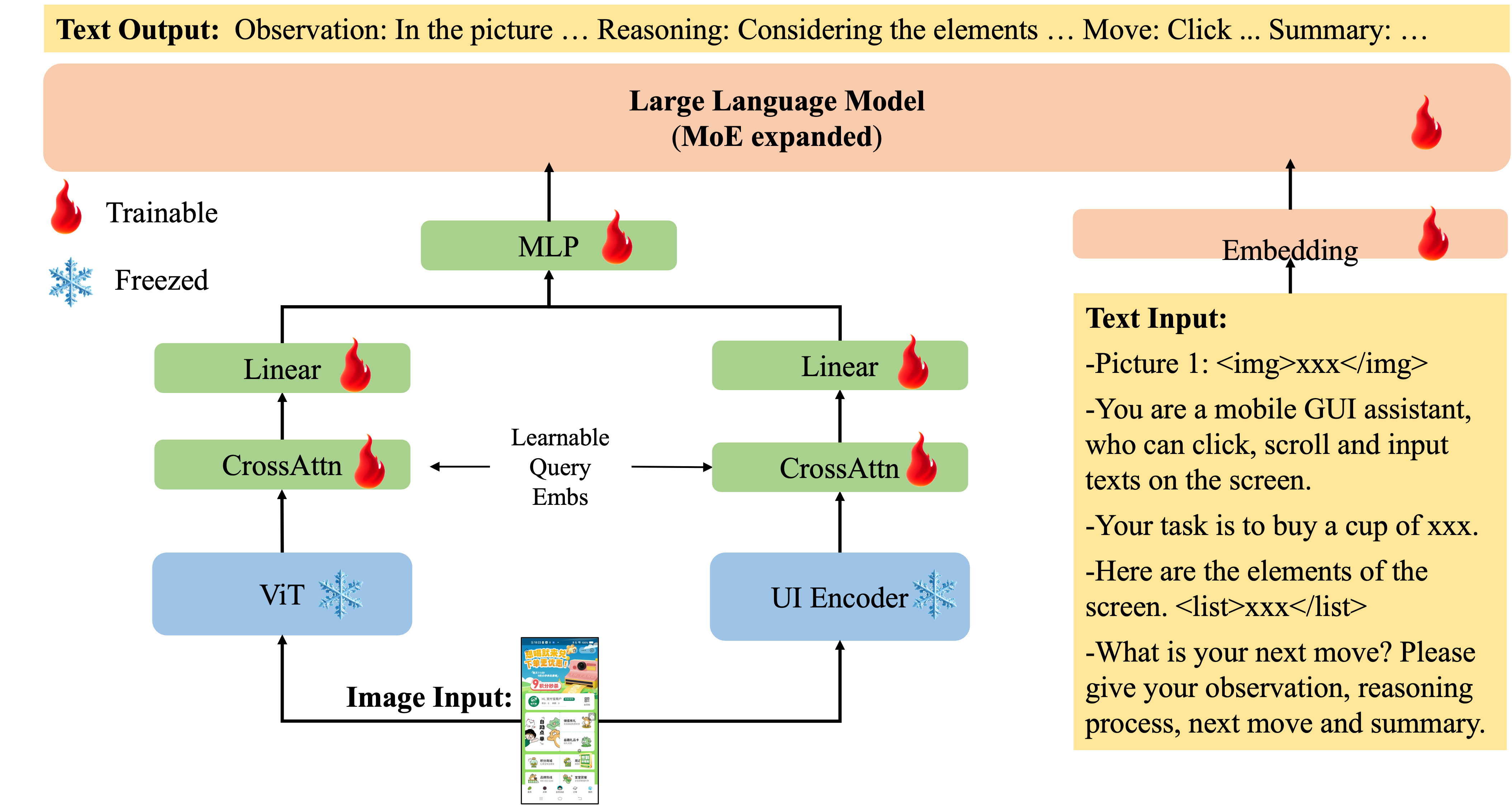}  
    \caption{Overview of MobileFlow.}  
    \label{fig:figure3}  
    \vspace{-2em}
\end{figure}

MobileFlow combines a visual encoder with a large language model through a fusion module for joint training with image-text pairs. It uses a Qwen-7B-based language model as a universal interface, paired with a visual perception module to gain dual "visualizing" capabilities. The paper's architectural framework has three main parts: the visual encoder, the visual-language adapter, and the extensive language model, as shown in Figure \ref{fig:figure3}. This section will detail the enhancements and optimizations made for GUI tasks on the original model structure.

\textbf{Hybrid Visual Encoders }
The architecture of the proposed visual encoder is illustrated in Fig.\ref{fig:figure4}. In line with most Vision-Language Models (VLMs), we construct our visual perception model based on the pre-trained Vision Transformer (ViT) structure. Specifically, for the ViT component, we utilize OpenAI's OpenCLIP ViT-B/32 pre-trained weights for initialisation. In addition, we introduce a UI Encoder, with capabilities for variable resolution input, to augment the extraction of visual information. After extensive research, we have chosen the document intelligence model LayoutLMv3, pre-trained with extensive document data, as the foundational structure for UI Encoder. We reassess and recalibrate the visual model's weights through redesigned UI image pre-training tasks on UI Encoder. In order to preserve the original aspect ratio of UI images to the greatest extent possible, we propose a variable resolution-based image encoding methodology, which is explained as follows.

When we set the target image sequence length for UI Encoder to be 784 tokens, with each image patch sized at 16x16 pixels, and the input image size is 1216x576, yielding an aspect ratio of 19:9, the resolution is recalculated by dynamically adjusting the width and height to compute an extreme aspect ratio while maintaining the original aspect ratio as closely as possible, under the sequence length constraint. In this example, the calculated number of image patches in the width and height directions are 41 and 19, respectively. Therefore, the recalculated dimensions for width and height are 41x16 and 19x16, respectively. And the total number of image patches amounts to 41x19=779, and the remaining 5 patches will be filled through padding.
\subsection{Model Architecture}

\begin{figure}[htbp]  
    \centering  
    \includegraphics[width=1.0\textwidth]{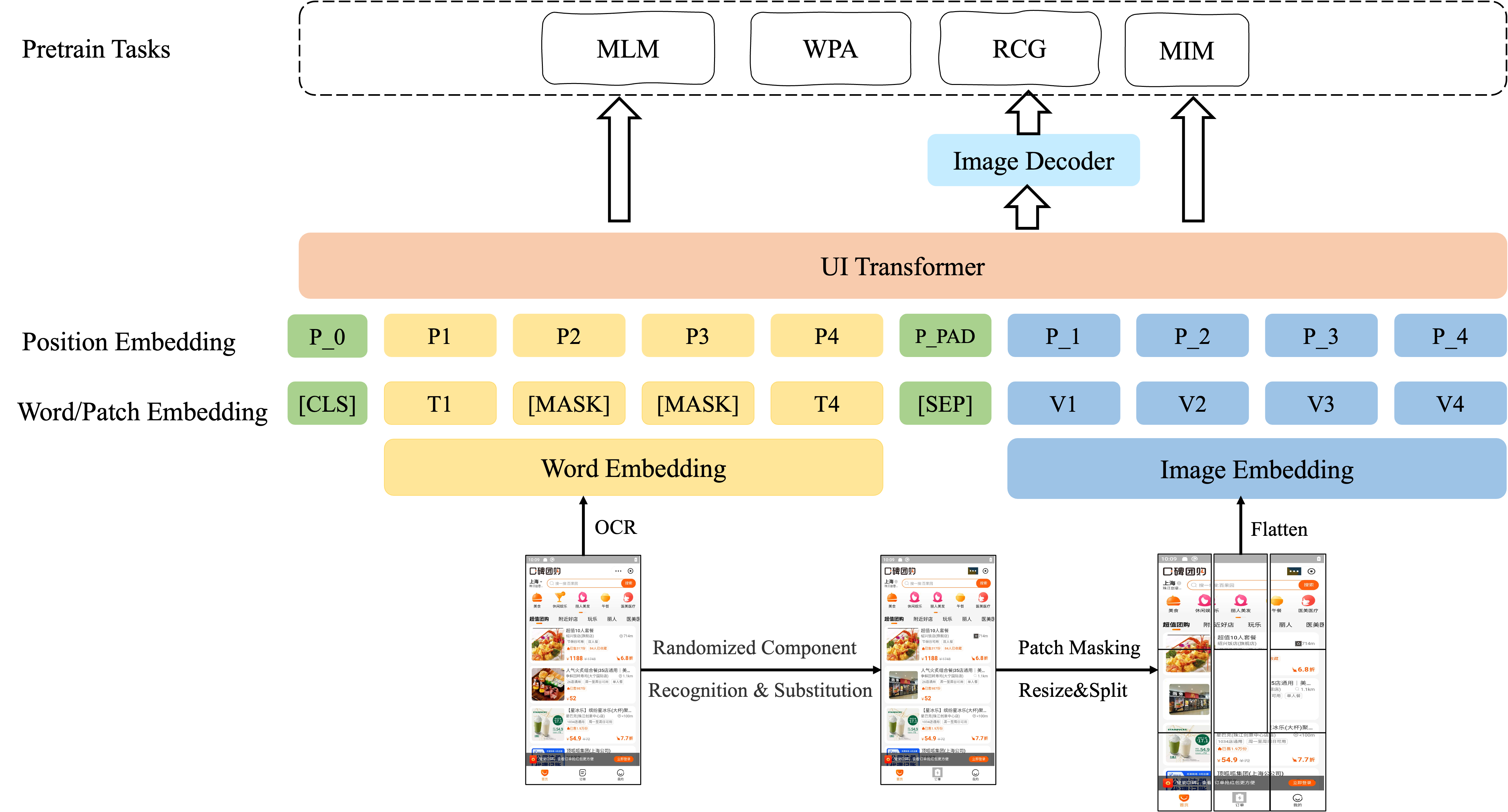}  
    \caption{Overview of UI Encoder.}  
    \label{fig:figure4}  
\end{figure}

As shown in Fig.\ref{fig:figure4}, we employ MLM (Masked Language Modeling), MIM (Masked Image Modeling), WPA (Word-Patch Alignment), and RCG (Real Component Generation) as pre-training tasks of UI Encoder. Both MLM and MIM are popular and widely used pre-training tasks, as referenced in 
 (\cite{devlin2019bertpretrainingdeepbidirectional, bao2022beitbertpretrainingimage}). The WPA task, introduced by the LayoutLMv3 paper (\cite{huang2022layoutlmv3pretrainingdocumentai}), predicts whether the corresponding image patch of a text word is masked, facilitating the learning of multi-modal alignment. The RCG task involves initially employing control recognition capabilities to detect all controls on a given UI interface, followed by randomly replacing these controls at a predetermined ratio. An image decoder is then utilized to reconstruct the original UI interface.

\textbf{Vision-Language Adapter}
MobileFlow introduces a Vision-Language Adapter designed to compress image features and fuse output features from multiple visual encoders. The adapter consists of the cross-attention mechanism and the MLP module. The cross-attention module employs a set of trainable vectors as query vectors, with image features from the visual encoders serving as key vectors, to condense each set of visual encoder features into a fixed length of 256. The MLP module integrates the features from parallel visual perception modules, projecting the visual features into the semantic space of the large language model with a minimal number of parameters. This configuration allows the model to flexibly perceive and understand visual modality information.

\textbf{MoE Expansion}
Many practices have shown that if LLMs adopt a Mixture of Experts (MoE) expansion, they can achieve a significant performance improvement while maintaining low inference costs. Most language models in current VLMs use a dense structure; hence, introducing the MoE approach to VLMs is also expected to provide considerable improvement. Generally speaking, there are two main methods of MoE expansion. One is to use random activations of multiple experts (where the routing is learnable, a typical example being Mixtral-8x7B (\cite{jiang2024mixtralexperts})), and the other is a combination of shared expert activations with random expert activations (with shared experts capturing global features, a typical example being DeepSeek-Chat (\cite{deepseekai2024deepseekllmscalingopensource})). In terms of implementation difficulty, this paper adopts the same method of random activations of multiple experts as Mixtral.

For MoE architecture models, an MoE layer typically contains multiple feedforward networks (FFNs). To leverage the visual understanding and dialogue question-answering capabilities learned during multi-stage training by Qwen-VL-Chat, MobileFlow adopts the method of directly duplicating the original MLP for expansion. Each MoE layer obtained after expansion includes 4 identical MLPs as initial experts. Multiple studies have shown that using trained MLPs as initial experts leads to quicker and more stable convergence during subsequent training and ultimately better performance compared to randomly initialized experts.

To sum up, MobileFlow's architecture consists of three main components, similar to other VLMs: a hybrid visual understanding network with multiple encoders, a visual-language alignment module, and an LLM enhanced with MoE. During training and inference, GUI screenshots are processed by the network to extract both global and detailed local features. These visual tokens, combined with text tokens from user inputs, OCR text, and BBOX data from the GUI, are fed into the LLM after alignment.
\subsection{Training Formulation}

\textbf{GUI Alignment}
For large multimodal models, training happens in two phases. First is visual-language alignment pre-training, where the model learns to connect images with text for appropriate responses. Second is instruction fine-tuning, where it learns to follow instructions for complex tasks like VQA, visual reasoning, and dialogue. Qwen-VL-Chat, trained on a lot of data, excels in image-text tasks after these phases. MobileFlow, inheriting Qwen-VL-Chat's multilingual image understanding, doesn't need full-scale pre-training but light GUI alignment and fine-tuning for good GUI agent skills. This light stage includes four training tasks:

\textit{GUI Grounding}: The purpose of this task is to help the model establish connections between text and specific areas in an image. Given that app pages typically contain rich textual information and various UI designs, incorporating this type of task can enhance the model's spatial understanding of the page.

\textit{GUI Referring}: Given specific bounding boxes or spatial references in text descriptions (such as upper left, lower right, etc.) or number references (first, last, third on the right, etc.), the model is required to output textual information at those positions. The model can understand the content referred to by the text and identify and locate the referent object in the image, which is crucial for a GUI Agent since many users’ actual intentions often include references.

\textit{UI Image Question Answering}: Here, we use the open-source ScreenQa dataset[28] to familiarize the model with and understand mobile GUI interfaces, transferring its foundational VQA capabilities to GUI-styled page content. The ScreenQA dataset contains a diverse range of VQA types, including both the extraction of page information and the inference of page content.

\textit{Image Description with Object Location}: MobileFlow is required to describe the image in details with the objects' bounding boxes. This task further reinforces the model's spatial understanding of GUI pages.

\textbf{GUI Chain-of-Thought}
The core idea of CoT is to have the model generate a series of intermediate steps or explanatory statements before delivering the final answer. These steps resemble the human thought process in problem-solving and gradually lead to the derivation of the solution. However, in most current multimodal LLMs, models tend to directly output answers without providing the reasoning process or rationale. This approach often fails in scenarios that require high-level logical reasoning (e.g., in a GUI Agent where continual decision-making, clicking, or swiping is necessary to fulfill user intentions). Therefore, in MobileFlow, we employ the CoT technique in both training and inference of the model. After being modified with CoT, the model shows a noticeable improvement in link accuracy and question-answering accuracy.MobileFlow adopts a Chain of Thought definition similar to AppAgent, where intent execution tasks consist of four steps:

    \textit{Observation}: Describing the contents observed on the GUI page, integrating information about page controls.
    
    \textit {Reasoning}: Considering how to operate on the current page to accomplish the given task.
    
    \textit {Action}: Generating behaviors within the Agent’s action space, which could be clicking, swiping, or typing text.
    
    \textit {Summary}: Summarizing the actions and previous behaviors as historical information for the next round interaction.

The task structure for visual question answering is similar, except that the action step is replaced with generating an answer. The detailed prompt structure is shown in Appendix B.

\section{Experiments}
In this section, we then show the qualitative and quantitative outcomes for the action prediction and visual question answering tasks, complemented by an ablation study that underscores the significance of our technical contributions. And more experiments implementation and deployment details are demonstrated in Appendix D.

\subsection{Metrics}
In prior research, MobileAgent introduced a set of metrics that were effective for discrete counting with a limited number of samples—specifically, the paper referenced a case with just 10 samples. However, when scaling up to larger test datasets, these metrics fail to accurately reflect the capabilities of the proposed Large Language Model (LLM) agent. Additionally, in response to the observed issue of endpoint determination, we introduced the Endpoint Determination Rate (EDR) as a new metric to identify this problem. We showed how to determine whether cases are positive or negative in Appendix C.

Furthermore, we have adopted the Whole Task Success Rate (WTSR) and Step Success Rate (SSR), metrics mentioned in previous studies, to assess the accuracy of both single-step and multi-step predictions. The calculations of each metrics are demonstrated in Appendix C.

\subsection{Quantitative Results}
To thoroughly assess the capabilities of our newly proposed method, we conducted a comparative analysis of MobileFlow against other leading Large Language Model (LLM)-based terminal agent algorithms, including MobileAgent and GPT-4v, in the context of mobile application strategy generation across various business domains.
\begin{table}[htbp]
    \centering
    \caption{Quantitative Results of MobileFlow in 6 business areas and 3 complexity of tasks}
    \label{tab:table1}
    \resizebox{\textwidth}{!}{
        \begin{tabular}{c|c|c|c|c|c|c|c|c}
        \hline
        Complexity & Metrics & Food Delivery & Food Walkin & Medical Service & Fund Select & Insurance & Gaming & All\\ 
        \hline
        \multirow{3}{*}{Long Chain Tasks} & WTSR & 0.2353 & 0.1765 & 0.1875 & 0.2857 & - & - & 0.2213 \\ 
        & SSR & 0.8282 & 0.7665 & 0.8163 & 0.5652 & - & - & 0.7441\\
        & EDR & 0.1429 & 0.14 & 0.1574 & 0.1038 & - & - & 0.1360\\
        \hline
        \multirow{3}{*}{Middle Chain Tasks} & WTSR & 0.7691 & 0.3999 & 0.5554 & 0.2308 & 0.2 & - & 0.4310 \\ 
        & SSR & 0.9634 & 0.9032 & 0.8947 & 0.5652 & 0.7547 & - & 0.8162 \\ 
        & EDR & 0.3241 & 0.423 & 0.1739 & 0.1796 & 0.1875 & - & 0.2576\\
        \hline
        \multirow{3}{*}{Short Chain Tasks} & WTSR & - & 0.9995 & - & 0.4154 & 0.7999 & 0.6363 & 0.7128\\ 
        & SSR & - & 0.9997 & - & 0.4386 & 0.8332 & 0.7199 & 0.7479 \\
        & EDR & - & 0.25 & - & 0.2015 & 0.2890 & 0.2047 & 0.2363\\
        \hline
        \multirow{3}{*}{Average} & WTSR & 0.4667 & 0.2917 & 0.3333 & 0.3810 & 0.3333 & 0.6363 & 0.4071 \\ 
        & SSR & 0.8735 & 0.7921 & 0.8341 & 0.5353 & 0.6136 & 0.7199 & 0.7280 \\ 
        & EDR & 0.3 & 0.2083 & 0.2593 & 0.1807 & 0.2450 & 0.2047 & 0.2330\\
        \hline
    \end{tabular}
    }
\end{table}

For our quantitative assessment, we divided the test dataset into six business sectors: Food Delivery, Walk-in, Insurance, Medical, Fund Selection, and Gaming Apps. We also categorized tasks into three complexity levels: long chain (over 8 steps), middle chain (4-8 steps), and short chain (4 steps or less). The comparative results are presented in Tab.\ref{tab:table1}, offering a detailed view of MobileFlow's performance metrics in comparison to existing state-of-the-art solutions.

Based on the data presented in Tab.\ref{tab:table1}, several key observations can be made. The performance across different business sectors varies significantly, which may be attributed to the differing distributions of task step lengths, or complexities, within each area. Additionally, there is a clear trend indicating that as task complexity increases, the performance of the evaluation metrics tends to degrade, a phenomenon that aligns with human cognitive patterns.

EDR may initially seem low, but it's tied to WTSR. A task must fully execute and end correctly for EDR to count it as successful. Thus, EDR is expected to be lower than WTSR. We found this to be true, especially in Medical Service Apps, where all tasks successfully predicted also ended properly.

Furthermore, we conducted a comparative analysis of our proposed MobileFlow against the current state-of-the-art algorithms. The comparative results are detailed in Tab.\ref{tab:table2}, offering insights into how MobileFlow stacks up against existing leading solutions in the field.
\begin{table}[htbp]
    \centering
    \caption{Comparison with current SOTA LLM-agent on action prediction task}
    \label{tab:table2}
    \resizebox{\textwidth}{!}{
        \begin{tabular}{c|c|c|c|c|c|c|c|c}
        \hline
        Method & Metrics & Food Delivery & Food Walkin & Medical Service & Fund Select & Insurance & Gaming & All\\ 
        \hline
        \multirow{2}{*}{GPT-4v} & WTSR & 0.1833 & 0.1876 & 0.1138 & 0.1428 & 0.2021 & 0.4132 & 0.2071 \\ 
        & SSR & 0.5716 & 0.4647 & 0.3571 & 0.4857 & 0.5012 & 0.6613 & 0.5069\\
        \hline
        \multirow{2}{*}{Qwen-VL-Max} & WTSR & 0.3650 & 0.2562 & 0.1643 & 0.2875 & 0.3076 & 0.5832 & 0.3273 \\ 
        & SSR & 0.7338 & 0.7075 & 0.6962 & 0.5112 & 0.2425 & 0.7763 & 0.6113 \\
        \hline
        \multirow{2}{*}{MobileFlow} & WTSR & 0.4667 & 0.2917 & 0.3333 & 0.3810 & 0.3333 & 0.6363 & 0.4071\\ 
        & SSR & 0.8735 & 0.7921 & 0.8341 & 0.5353 & 0.6136 & 0.7199 & 0.7280 \\
        \hline
    \end{tabular}
    }
\end{table}

Comparing MobileFlow to other top LLM agents shows it has better performance across different sectors. MobileFlow excels in complex tasks. Despite having a smaller LLM than Qwen-vl-max, MobileFlow's results are competitive, as shown in Table \ref{tab:table2}. This indicates that even with a smaller model, MobileFlow can improve after fine-tuning and can match or outperform larger models, proving the method's efficiency and strength.

For the visual-question-answering tasks, the results are presented in the subsequent Tab.\ref{tab:table3}. Table 3 offers a breakdown of how MobileFlow and other leading LLM agents perform on tasks that involve interpreting both visual and textual data to produce precise responses. It assesses their capabilities in visual-question-answering, focusing on the accuracy of the answers and the ability to understand and respond correctly to questions presented with visual cues. Our MobileFlow has shown strong VQA abilities, especially after fine-tuning with business-specific data. It's impressive because it can stand up to larger models like Qwen-vl-max. This shows that our fine-tuning works well and that MobileFlow can do great in VQA tasks even if it's not as big.

\begin{table}[htbp]
    \centering
    \caption{Comparison with SOTA LLM-agent on VQA task}
    \label{tab:table3}
    {\small
        \begin{tabular}{c|c|c|c}
        \hline
        Method & Recall & Accuracy & F-score \\ 
        \hline
        GPT-4v & 0.6835 & 0.6228 & 0.6521 \\
        \hline
        Qwen-VL-Max & 0.7478 & 0.7064 & 0.7268\\
        \hline
        MobileFlow & 0.7478 & 0.7253 & 0.7363 \\
        \hline
        \end{tabular}
    }
\end{table}

\subsection{Ablation Study}
\textbf{The Effectiveness of MoE structure}
Based on the data from Tab.\ref{tab:table4}, the newly proposed architecture of ViT (Vision Transformer) with MoE (Mixture of Experts) in MobileFlow has achieved significant improvements over the conventional ViT plus dense Large Language Model (LLM) structure. Specifically, there is an 4.49\% enhancement in Whole Task Success Rate (WTSR) and a notable 8.17\% increase in Step Success Rate (SSR). These results underscore the effectiveness of the MoE components in enhancing the performance of MobileFlow, highlighting the benefits of this innovative model structure for complex task execution and prediction accuracy.

\textbf{The Effectiveness of the UI encoder}
We compared the performance of MobileFlow before and after the inclusion of the UI Encoder. Experimental results indicate that incorporating the UI Encoder as a visual branch into the MobileFlow architecture resulted in a 6.96\% improvement in the WTSR metric and a 4.47\% improvement in the SSR metric. This further underscores the importance of supplementing UI visual information for the final decision-making in MobileFlow.

\begin{table}[htbp]
    \centering
    \caption{Effectiveness of the MoE structure and the UI encoder}
    \label{tab:table4}
    {\small
        \begin{tabular}{c|c|c}
        \hline
        Model & WTSR & SSR \\ 
        \hline
        ViT(448px) + Dense & 0.2926 & 0.6016 \\
        \hline
        ViT(448px) + MoE & 0.3375 & 0.6833\\
        \hline
        ViT(448px) + UI Encoder + MoE & 0.4071 & 0.7280\\
        \hline
        \end{tabular}
    }
\end{table}

\section{Conclusion}
In this paper, we present MobileFlow, a GUI Agent that leverages a multimodal large model and incorporates a set of optimization techniques, to analyze UI images, understand user instructions and operate under the practical scenarios. Our proposed MobileFlow has been designed to navigate a diverse array of intricate scenarios and business domains with proficiency. It has demonstrated its capabilities in practical applications across various sectors, showcasing its versatility and effectiveness. And more applications could adopt MobileFlow for further optimization as shown in Appendix D. 

As with many pioneering agents in the industry, MobileFlow marks an important initial step in the evolution of GUI Agents. MobileFlow faces future challenges like susceptibility to hallucinations and  multiple images handling. It's mainly used for mobile apps now but has potential to expand to other devices like computers. This could turn MobileFlow into a reliable, user-friendly AI assistant for everyday tasks across platforms.


\bibliographystyle{unsrtnat}
\bibliography{sample}  





\clearpage
\appendix
\section{Action Space}
A GUI Agent requires ongoing interaction with the Graphical User Interface (GUI) to accomplish tasks set forth by human. An interaction can be interpreted as either a singular action or an amalgamation of multiple actions. Hence, the judicious design of the action space is crucial for enhancing the effect of a GUI Agent. An overly simplistic action space could limit the variety of tasks that the GUI Agent is capable of executing. Thus, it becomes imperative to devise an elaborate action space capable of encompassing the majority of tasks within mobile GUI contexts. As illustrated in Tab.\ref{tab:table5}, based on actual usage requirements, we have designed a total of 8 actions and provided detailed explanations for each.
\begin{table}[htbp]
    \centering
    \caption{Action Space}
    \label{tab:table5}
    {\small
        \begin{tabular}{c|c|c}
        \hline
        Action & Parameters & Explanation \\ 
        \hline
        Click & Position & Click at a specified location \\
        \hline
        Long Press & Position & Long press at a specified location\\
        \hline
        Input & Text & Input the text at the current cursor position\\
        \hline
        Scroll & Position List & Slide along a list of positions\\
        \hline
        Drag & Position List & Long press, then slide along a list of positions\\
        \hline
        Wait & Time & Wait without performing any actions\\
        \hline
        Task Finish & - & Upon completion of the current task, the agent ceases operation\\
        \hline
        \end{tabular}
    }
\end{table}
\section{MobileFlow Details}
\subsection{Prompt structure}
The ultimate detailed prompt structure is as follows:
\begin{verbatim}
Picture 1: <img> image_path </img>
Imagine you are a Mobile GUI assitant. Just like a human operating a mobile phone, you 
can tap and swipe the page, or use the keyboard to type some text, and you can also 
answer questions.

Your task is <task> task info </task>.
The elements on the page are as follows. <list> elements </list>
Your historical moves to advance this mission are summarized below. <history> ... 
<history>

Based on your task, historical actions, and the current page information, you need to 
think and generate actions that can advance the task. You should respond in the 
following format:
<observation>: Based on control information, describe the contents observed on the page.
<Reasoning>: To accomplish the task, contemplate what action should be generated next.
<Action>: A feasible action to accomplish the task, which could be a click, swipe, 
or input text. When you determine that the task is completed, you may also output 
"Finish".
<Summary>: Assuming the next action has been executed, summarizing in two or three 
sentences in conjunction with the history of actions performed thus far, without 
including any potential future actions or specific coordinates of controls.
\end{verbatim}

\section{Evaluation Details}
\subsection{Positive Sample Determination}
\begin{figure}[htbp]  
    \centering  
    \includegraphics[width=1.0\textwidth]{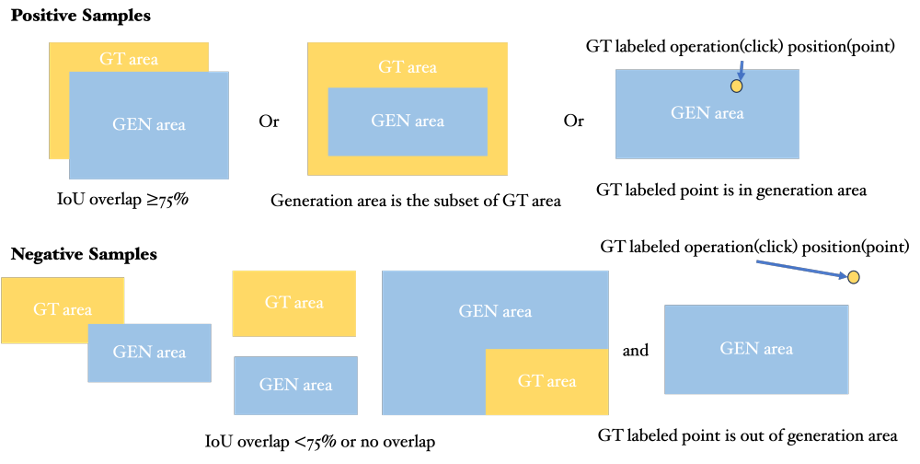}  
    \caption{Matching cases for positive and negative samples.}  
    \label{fig:figure5}  
\end{figure}
Determining the correctness of a prediction is not always a clear-cut or binary decision. To address this, our proposed metrics employ a combination of methods: matching the predicted action type and calculating the Intersection over Union (IoU) of the coordinate areas to assign a true or false label to each prediction. This approach allows for a more nuanced evaluation that accounts for the complexity of the tasks and the subtleties of the predictions made by the LLM-agent. Specifically, we illustrate all the matching cases in Fig.\ref{fig:figure5}, providing a visual representation of how these metrics are applied to assess the accuracy of predictions.

\subsection{Deployment Details}
\subsection{Dataset}
To fully train the MobileFlow pipeline, we have specifically trained two models: UI Encoder for user-interface understanding and Qwen-vl-chat for token prediction.

For the training of UI Encoder, we meticulously curated and cleaned a dataset of 100K manually labeled instances that are directly relevant to our current business domains. This dataset was then used to fine-tune UI Encoder

For the multimodal alignment, we employed a blend of the RefCoCo[29], ScreenQa[28], Flickr30K[9] and in-house UI datasets. Subsequently, for supervised fine-tuning, we utilized 70k manually labeled business-specific data. These data were collected across 10 distinct business sectors, such as food delivery applications, medical service platforms, insurance applications, financial applications, and more. In constructing these data, we concurrently devised a comprehensive action space, details of which can be found in Appendix A.

\subsection{Training and Evaluation details}
In our experiments, we pre-train UI Encoder with approximately 200k UI images, improving the model's comprehension of fonts, images, controls, and other elements within UI imagery, building upon its fundamental ability to understand documents. The effectiveness of the pre-training tasks was specifically validated through multiple UI image downstream tasks, such as image classification and ui component recognition.

During the training of UI Encoder, we conducted fine-tuning on our 100K labeled business dataset over a span of 2 epochs. For the supervised fine-tuning phase of Qwen-vl-chat, after closely monitoring the loss function outcomes, we determined that training it for 2 epochs with a learning rate of yielded the optimal performance. For the evaluation phase, we employed MobileAgent, making necessary adjustments to the data format and interface to ensure compatibility with our test data, all the while keeping the model parameters intact. Additionally, we directly utilized the GPT-4v interface for token prediction on our test data. All experiments were conducted on a robust setup of 8 GPUs (Nvidia A100).

\subsection{Definition of Metrics}
\textbf{Whole Task Success Rate(WTSR)}
The Whole Task Success Rate (WTSR) is a metric that measures the proportion of instances where the Large Language Model (LLM) agent can successfully predict every step within a single task across the entire dataset. 
\begin{equation}
    WTSR=\frac{\#SuccessIntentions}{\#AllIntentions - \#TimeOut} 
\label{eq:WTSR}
\end{equation}

\textbf{Step Success Rate(SSR)}
Given a specific intention, the Step Success Rate (SSR) measures the frequency at which the Large Language Model (LLM) agent can accurately predict each individual step within all multi-step tasks. 
\begin{equation}
    SSR=\frac{\#SuccessSteps}{\#AllSteps} 
\label{eq:SSR}
\end{equation}

\textbf{Endpoint Determination Rate(EDR)}
The Endpoint Determination Rate (EDR) is a metric that quantifies the proportion of tasks that the Large Language Model (LLM) agent successfully concludes on the final page of the entire dataset.
\begin{equation}
    EDR=\frac{\#SuccessTerminalIntentions}{\#AllIntentions - \#TimeOut} 
\label{eq:EDR}
\end{equation}

\section{Applications}
\subsection{Software Testing}
Software testing is an ideal scenario for MobileFlow. First of all, software testing requires significant amount of work to complete. Based on internal survey, software testing in average takes 150\% longer compare to development in time. In practice, testing account status, testing data, pop-ups, algo driven UI displays, a/b tests may create UI route noises in traditional automation executions, causing false alarms. The success rate is usually low. Automation script frequently require updates to be compatible to thses noise in UI routes.

MobileFlow solves this issue from three aspects:
\begin{itemize}
    \item Replacing test automation script by natual language reduces the complexity of testing system and programming skill prerequisite for testing engineers.
    \item Natual language has good capacity to deal with UI noises, and sometimes can be compatible to testing account/data differences.
    \item Some alarms can be automatically analyzed and closed by VAQ task.
\end{itemize}

\subsection{Advertisement Preview And Audit}
Advertisement contains multimedia contents and requires muti-step interactions to trigger. This cause manual and traditional automation are costly to execute and unstable in success rate.

In this scenario, MobileFlow can serve from both advertiser and advertising platform perspectives.
\begin{itemize}
    \item Advertiser: Trigger ads, preview the ads as expected.
    \item Advertising platform: Monitoring ads trigger stragegy and interaction logic, audit ads content to prevent improper content to display.
\end{itemize}

\subsection{E-Commerce Operation And Monitoring}
Most of small merchants in China have their E-Commerce business and they are operated on different platforms, such as Taobao store, Jingdong store, PDD store, Alipay mini app, WeChat mini app, Tiktok shop, Meituan shop, RED shop, etc. Daily marketing and operation are a burden for business owners, and small mistakes may cause big loss, even lead to bankruptcy.

A small tool based on MobileFlow is developed for small merchants to perform the following tasks:
\begin{itemize}
    \item Inspecting price discrepencies cross platforms, usually caused by mistakes in coupon/discount setup.
    \item Inspecting price discrepencies cross dates, usually caused by platform level marketing event.
    \item Monitoring competitive stores marketing events and follow-up.
\end{itemize}

\section{More Samples}
\begin{figure}[htbp]  
    \centering  
    \includegraphics[width=1.0\textwidth]{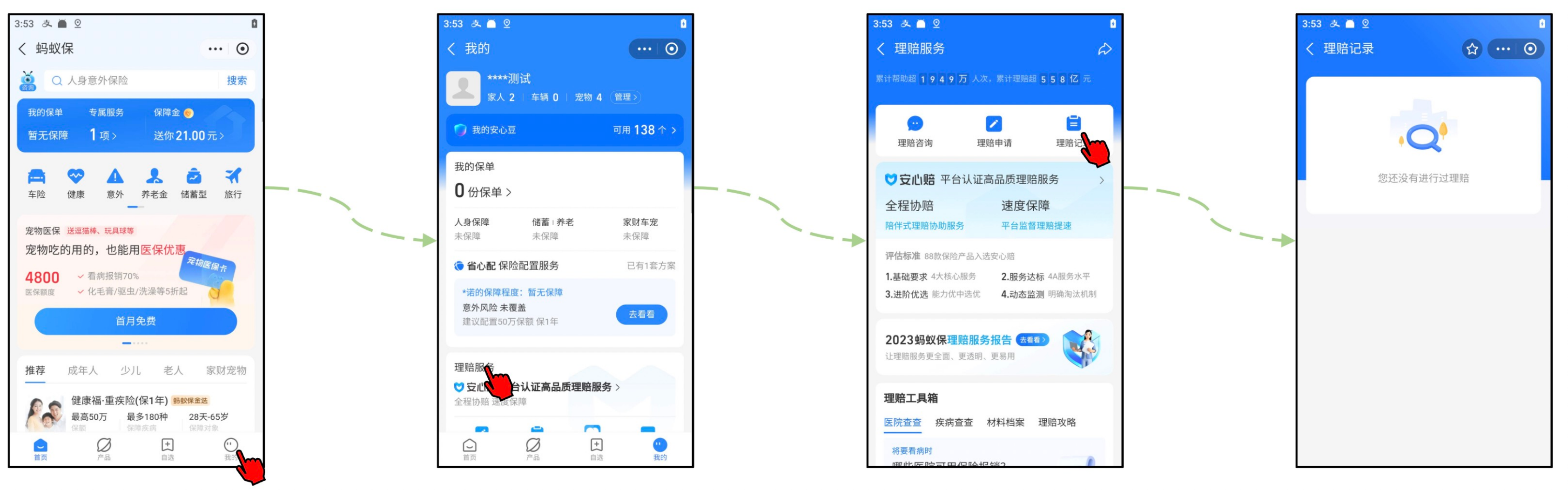}  
    \caption{User's instruction: Check my claim records.}  
    \label{fig:figure7}  
\end{figure}

\begin{figure}[htbp]  
    \centering  
    \includegraphics[width=1.0\textwidth]{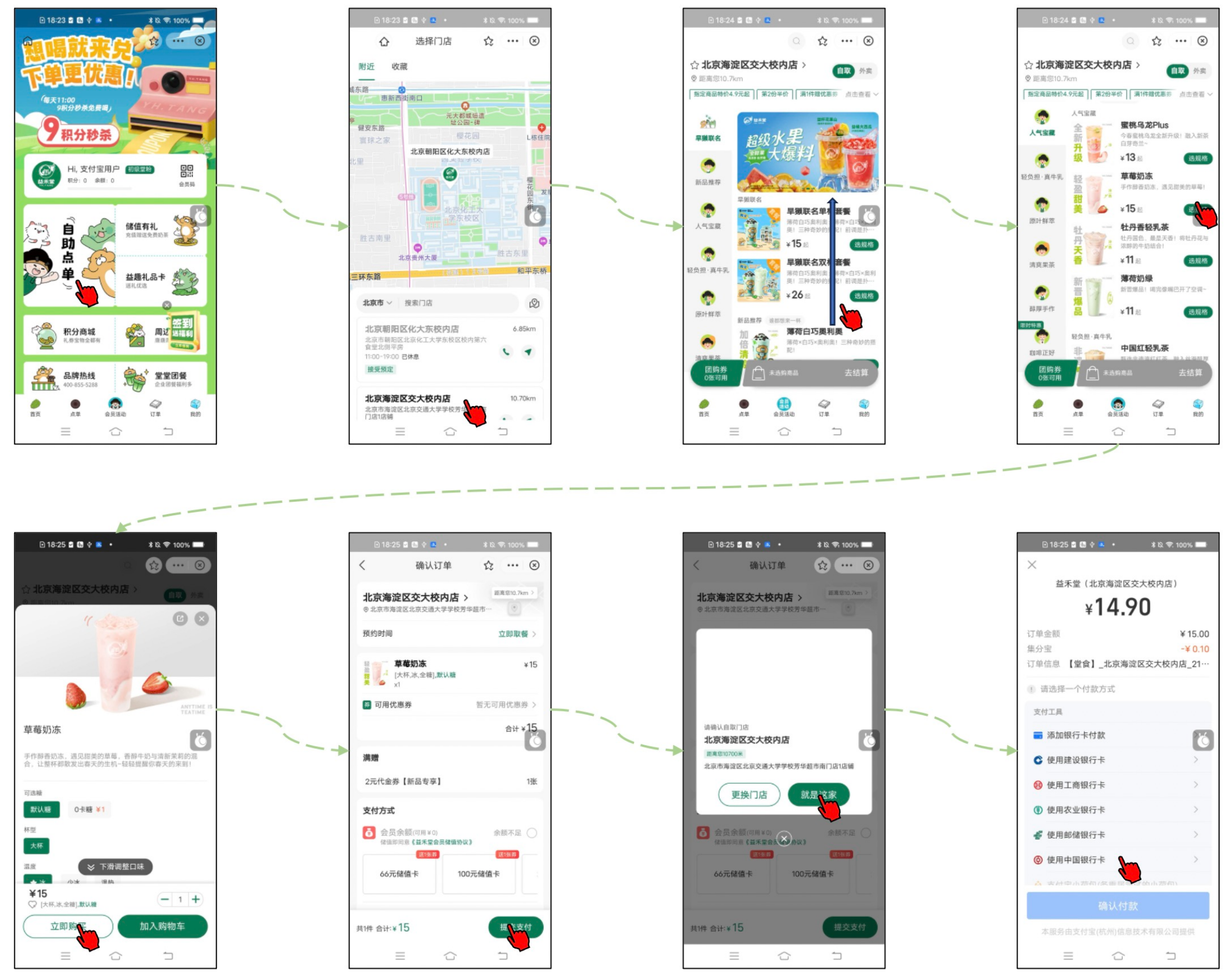}  
    \caption{Showcase of the MobileFlow's application for GUI agent. User's instruction: Help me buy a cup of strawberry custard in the self-service order, choose Jiao Tong University Campus store of Haidian District.}  
    \label{fig:figure1}  
\end{figure}
\begin{figure}[htbp]  
    \centering  
    \includegraphics[width=1.0\textwidth]{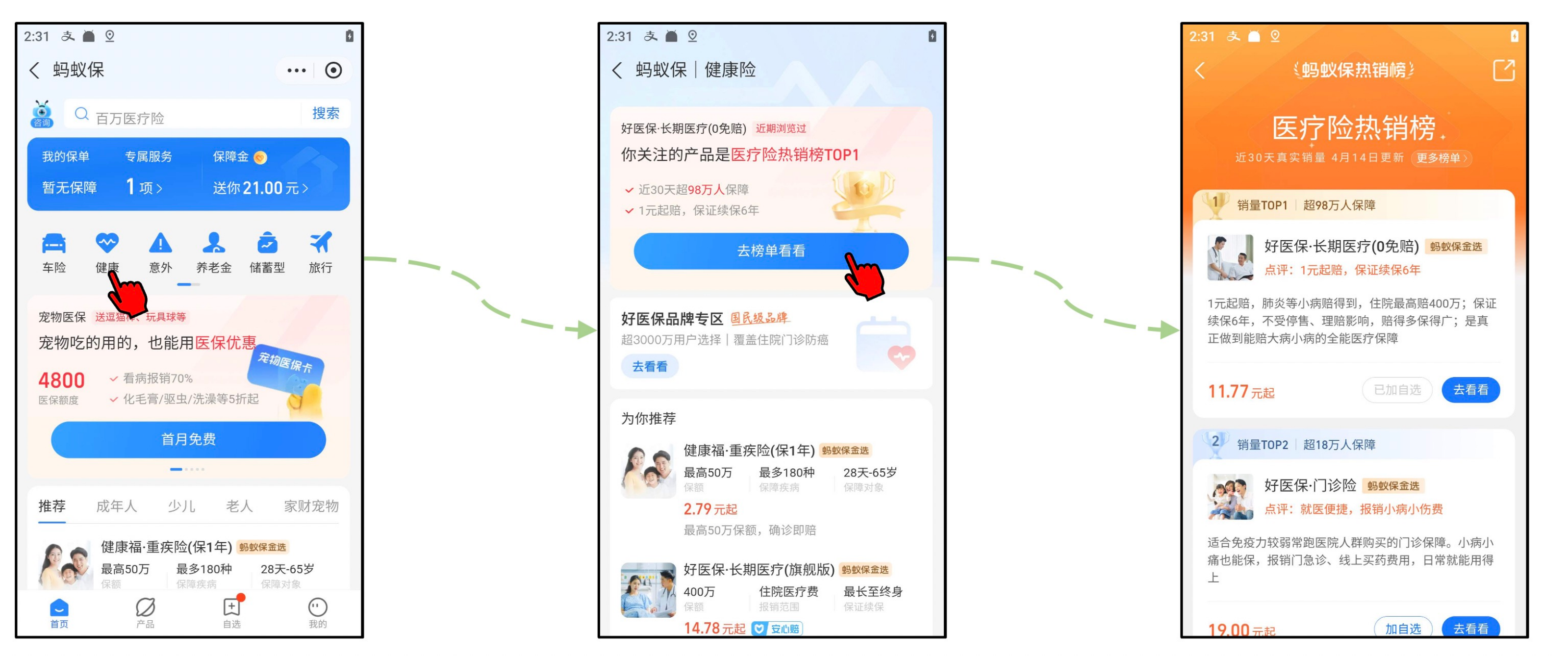}  
    \caption{User's instruction: Check out the top health insurance lists.}  
    \label{fig:figure6}  
\end{figure}

\begin{figure}[htbp]  
    \centering  
    \includegraphics[width=1.0\textwidth]{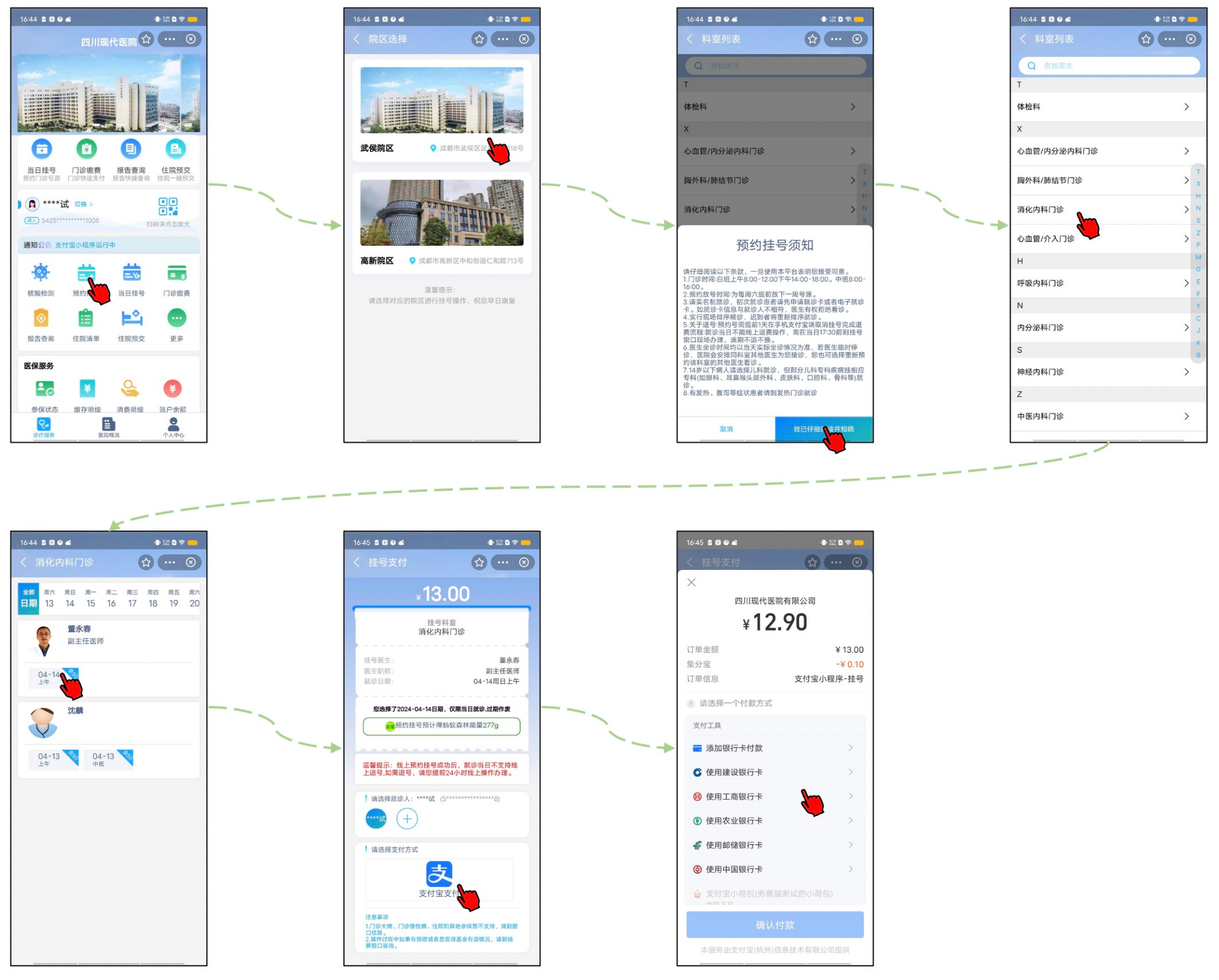}  
    \caption{User's instruction: Please make an appointment for the gastroenterology clinic on April 14th. Choose Wuhou Hospital of Sichuan Modern Hospital.}  
    \label{fig:figure8}  
\end{figure}

\end{document}